%% file: acl.tex
\PassOptionsToPackage{table,xcdraw}{xcolor}  
\documentclass[11pt]{article}

\usepackage[final]{acl}

\usepackage{times}
\usepackage{latexsym}

\usepackage[T1]{fontenc}

\usepackage[utf8]{inputenc}

\usepackage{microtype}

\usepackage{inconsolata}

\usepackage{graphicx}
\usepackage{amsmath}   
\usepackage{amssymb}   
\usepackage{graphicx}  
\usepackage{graphicx} 
\usepackage{wrapfig}  
\usepackage{lipsum}   
\usepackage{float}    
\usepackage{amsmath}
\usepackage{amssymb}
\usepackage{tcolorbox}
\usepackage{hyperref}  
\usepackage{multirow}
\usepackage{graphicx}
\usepackage[normalem]{ulem}
\useunder{\uline}{\ul}{}
\usepackage{array}
\usepackage{booktabs}
\usepackage{mathtools}
\usepackage{xcolor}

\tcbset{
    mybox/.style={
        colback=gray!10, 
        colframe=black, 
        coltitle=white, 
        fonttitle=\bfseries, 
        boxrule=0.75mm, 
        arc=1mm, 
        leftrule=2mm, 
        width=\linewidth, 
        boxsep=5pt, 
        left=10pt, 
        right=10pt, 
    }
}

\title{\textsc{Reward Consistency:} Improving Multi-Objective Alignment \\from a Data-Centric Perspective }

\author{
  Zhihao Xu$^{1 \thanks{Work done during an internship at Ant Group.}}$, 
  Yongqi Tong$^2$, Xin Zhang$^2$, Jun Zhou$^2$, 
  Xiting Wang$^{1 \thanks{{Corresponding author.}}}$ \\
  \\
  $^1$ Renmin University of China $^2$ Ant Group
  }  

\begin{document}
\maketitle

\begin{abstract}

Multi-objective preference alignment in language models often encounters a challenging trade-off: optimizing for one human preference (e.g., helpfulness) frequently compromises others (e.g., harmlessness) due to the inherent conflicts between competing objectives. While prior work mainly focuses on algorithmic solutions, we explore a novel data-driven approach to uncover the types of data that can effectively mitigate these conflicts. Specifically, we propose the concept of \textsc{Reward Consistency (RC)}, which identifies samples that align with multiple preference objectives, thereby reducing conflicts during training. Through gradient-based analysis, we demonstrate that RC-compliant samples inherently constrain performance degradation during multi-objective optimization. Building on these insights, we further develop \textsc{Reward Consistency Sampling}, a framework that automatically constructs preference datasets that effectively mitigate conflicts during multi-objective alignment. Our generated data achieves an average improvement of 13.37\% in both the harmless rate and helpfulness win rate when optimizing harmlessness and helpfulness, and can consistently resolve conflicts in varying multi-objective scenarios.



\end{abstract}

\input{main/1_intro}

\input{main/2_problem}
\input{main/3_method}
\input{main/4_framework}

\input{main/5_experiments}

\input{main/6_related_work}
\input{main/7_conclusion}


\bibliography{reference.bib}

\appendix
\input{main/appendix}

\end{document}

%% file: main/1_intro.tex
\section{Introduction}

Alignment is a critical stage in the fine-tuning of language models, designed to ensure that the generated responses align with human preferences and values~\cite{guo2025deepseek,lambert2024t,xu2024uncovering}.

While current Reinforcement Learning with Human Feedback (RLHF)~\cite{ouyang2022training} and direct preference alignment methods~\cite{rafailov2024direct,azar2024general,hong2024orpo, ethayarajh2024kto, meng2025simpo} have been proven effective for improving the general quality of generated responses, they still face the significant challenge of aligning with diverse and often conflicting human preferences~\cite{casper2023open, rame2024rewarded}. The inherent conflicts between different human preferences often lead to trade-offs~\cite{bai2022constitutional, lou2024spo}, where optimizing for one preference may degrade performance in another preference, hindering universal performance improvements across diverse alignment dimensions, which we refer to as {alignment conflict} in this paper. 

Recent advancements in multi-objective direct preference alignment have introduced algorithmic improvements to reduce optimization conflicts while avoiding the high cost and instability of the RLHF process~\cite{zhou2024beyond}.
For example, MODPO~\citep{zhou2024beyond} and SPO~\citep{lou2024spo} extend DPO by introducing a margin loss term into the objective function, thereby establishing a multi-objective-driven training process that ensures simultaneous optimization across competing objectives. However, their effectiveness is still inherently constrained by the data itself. If the data lacks inherent multi-objective alignment potential, algorithmic adjustments alone struggle to resolve conflicts between objectives. 

While data selection is important for preference alignment, constructing datasets that inherently balance multiple conflicting objectives remains fundamentally challenging. Existing data selection for alignment frameworks usually focus on how to enhance the performance of homogeneous preferences or specific tasks~\cite{khaki2024rs,pattnaik2024curry,lai2024step,cui2023ultrafeedback,wang2024helpsteer2}, but lacks exploration of data that balances multiple preference objectives. Therefore, it remains challenging to clearly understand the desirable properties of multi-objective data, as well as to determine effective ways to identify such data.

This crucial gap prompts our central investigation in the context of direct preference alignment: How can we effectively construct data that reduces conflicts between competing preference objectives for training?
By identifying and understanding the mechanisms driving alignment conflicts, we present \textsc{Reward Consistency} (RC), a desirable property suitable for multi-objective alignment.
Through gradient-based analysis, we establish that reward-consistent samples inherently preserve multiple objectives during optimization through constrained gradient divergence. 
We further propose \textsc{Reward Consistency Sampling (RCS)} framework, which first samples diverse candidate responses from LLMs for each input prompt, then applies reward consistency principle to filter out those conflicting ones. Our framework works well with both implicit and explicit reward signals, and we can selectively keep rewards consistent along specific dimensions for flexible control. The generated data is also compatible with different direct preference alignment algorithms. 
Overall, we make the following contributions:

\begin{itemize}
    \item We introduce the principle of \textsc{Reward Consistency (RC)} and demonstrate that samples satisfying this principle effectively mitigates conflicts between competing objectives from both theoretical and empirical aspects.\looseness=-1
    \item We propose, to the best of our knowledge, the first data-centric framework for multi-objective direct preference optimization called \textsc{Reward Consistency Sampling (RCS)}. This approach integrates reward consistency with sampling to reconstruct preference datasets that can help mitigate conflicts.
    \item We validate the proposed RCS framework through extensive experiments. For instance, when optimizing the helpfulness and harmlessness objectives, training on data constructed by RCS achieved an average performance improvement of 13.37\% in both harmless rate and helpfulness win rate compared to using the original dataset.
\end{itemize}

%% file: main/2_problem.tex
\section{Problem Formulation}

In this work, we construct data for multi-objective direct alignment methods~\cite{zhou2024beyond,lou2024spo}, which train language models through closed-form loss functions like DPO~\cite{rafailov2024direct}. These methods bypass explicit reward modeling and leverage offline pair-wise preference data to capture multiple human preferences. Compared with online reinforcement learning methods like Multi-Objective RLHF (MORLHF)~\citep{rame2024rewarded, dai2023safe}, direct alignment methods require fewer computing resources and significantly reduce costs. 

Existing multi-objective direct alignment pipelines typically train language models sequentially on specialized preference datasets $\{D_1,...,D_k \}$, where $\mathcal{K}$ denotes the total number of preference objectives and each preference dataset $D_i$ targets at aligning the $i$-th objective~\cite{lou2024spo}.
Since datasets for different objectives are constructed without considering other objectives, conflicts can be easily introduced, making the datasets suboptimal for multi-objective alignment tasks. More specifically, each dataset $D_i=\{(x^j, y^j_w, y^j_l)\}_{j=1}^{M}$, where $x^j$ denotes a user input prompt, $y^j_w$ denotes the corresponding winning (chosen) response, $y^j_l$ denotes the losing (rejected) response, and $M$ represents the number of samples in $D_k$. Here, response $y^j_w$ is only guaranteed to win response $y^j_l$ in terms of the $i$-th objective (e.g., helpfulness), and may be worse than $y^j_l$ in terms of other objectives (e.g., safety), thus introducing potential conflicts. 

To address this challenge, we aim to automatically construct new preference datasets that are specifically designed to support multi-objective alignment and can be used in any sequential training framework as described above. Specifically, given $\mathcal{K}$ preference objectives, our goal is to generate datasets that mitigate alignment conflict, a phenomenon where optimizing the current objective degrades the performance of previous objectives during the alignment process. This can be formulated as:

\textbf{Input:} $\mathcal{K}$ preference objectives and preference dataset $D_{j}=\{(x^i, y^i_w, y^i_l)\}_{i=1}^{M}$ for current preference objective $j \in \{1, ...,\mathcal{K}\}$, where $M$ represents the number of samples in $D_j$.

\textbf{Output:} Preference datasets $\{D_1, D'_2, ... D'_K\}$, each $D'_i$ reduces conflicts with previously trained objectives, thereby facilitating multi-objective sequential alignment. We do not change $D_1$ here since no conflict is introduced when there is only one objective.

%% file: main/3_method.tex
\section{\textsc{Reward Consistency}} 
\label{sec: rc}
In this section, we discuss the desirable properties that samples should possess to resolve conflicts in multi-objective alignment. To this end, 
we first define \textsc{Reward consistency} as the desirable property (Section~\ref{sec:define_rc}) and then demonstrate its utility in resolving conflicts through theoretical analysis (Section~\ref{sec:rc_analysis}) and
empirical experiments (Section~\ref{sec:rc_help}).

\subsection{Definition of \textsc{Reward Consistency}}
\label{sec:define_rc}
To resolve conflicts, we first identify the desirable property for samples. Our intuition is that if a winning response $y_w$ outperforms the losing response $y_l$ only in a subset of objectives but underperforms in others, optimizing based on this response pair may lead to a performance decline in the latter objectives. Accordingly, we define the concept of reward consistency as follows:

\noindent\textbf{Definition 1 \textsc{(Reward Consistency)}.} A sample $(x, y_w, y_l)$ is said to satisfy reward consistency if $y_w$ consistently receives a higher reward than $y_l$ across all $\mathcal{K}$ objectives: $    r_j(x,y_w) > r_j(x,y_l), \quad \forall j \in \{1, 2, \dots, \mathcal{K}\}$.

Existing datasets for multi-objective direct alignment contain a considerable amount of samples that do not satisfy reward consistency, since the dataset for one objective is constructed independently without taking other objectives into account. 
Take the commonly-used helpfulness preference dataset HelpSteer2~\cite{wang2024helpsteer2} as an example. In 40\% of its response pairs, the winning response fails to outperform the losing one in terms of harmfulness, thus optimizing by using this dataset may lead to a significant decrease in harmfulness. 

\subsection{Theoretical Analysis}
\label{sec:rc_analysis}

\begin{figure}[]
    \centering
\includegraphics[width=0.5\textwidth]{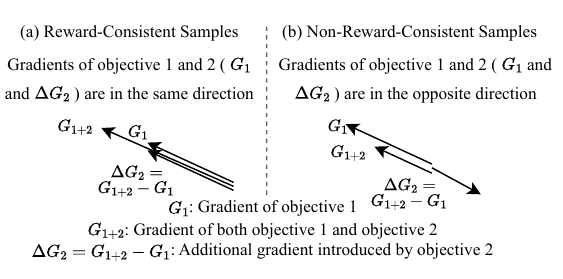}
\vspace{-8mm}
    \caption{Gradient analysis of reward consistency.}
    \label{fig:gradient}
\vspace{-3mm}
\end{figure}

To theoretically show how reward consistency resolves conflicts, we compare the gradients of reward consistent samples with samples that do not satisfy this property.  
Without losing of generality, we consider the scenario in which $\mathcal{K}=2$. 
Our observation is that for existing multi-objective direct alignment methods~\cite{zhou2024beyond,lou2024spo}, the gradient of the two objectives are not in the opposite direction (i.e., conflicting) if and only if the sample is reward consistent, as shown in Figure~\ref{fig:gradient}.
Formally, we have the following lemma:

\textbf{Lemma 1.} \textit{Let $\mathbf{G_1}$ represent the gradient of the current objective 1, $\mathbf{G_{1+2}}$ represent the gradient considering both objectives 1 and 2, and $\mathbf{\Delta{G_{2}}} = \mathbf{G_{1+2}} - \mathbf{G_1}$ denote the additional gradient introduced by considering objective $2$. 
$\mathbf{G_1} \cdot \mathbf{\Delta{G_{2}}} \geq 0$ (i.e., not conflicting with each other) in existing multi-objective direct alignment methods~\cite{zhou2024beyond,lou2024spo} if and only if the sample $(x, y_w, y_l)$ is reward-consistent.
} 

See Appendix~\ref{app:derivation} for detailed proof.
This analysis highlights the importance of reward consistency in reducing conflicts in multi-objective preference alignment.

\subsection{Empirical Experiments}
\label{sec:rc_help}
We now empirically validate that reward consistency can reduce conflicts during training. Table~\ref{tab:rc} shows the alignment performance when training with the original dataset for optimizing helpfulness,  the reward inconsistent samples in the dataset, and the reward consistent samples in the original dataset.
Results show that only reward consistent samples (RC) can ensure improvement on both harmfulness and helpfulness.
In contrast, training on reward inconsistent samples (NRC) or the original dataset (Org.) leads to \colorbox{red!10}{significantly degradation} in the harmless rate. This shows that reward consistency serves as an effective guiding principle for reducing conflicts between competing objectives during optimization.
More details of the experiment setups can be found in Appendix~\ref{app:details_rc}.

\begin{table}[htbp]
\resizebox{0.5\textwidth}{!}{%
\begin{tabular}{ccccc}
\hline
\textbf{\begin{tabular}[c]{@{}c@{}}
\end{tabular}} &
  \textbf{\begin{tabular}[c]{@{}c@{}}Harmless \\ Rate ↑\end{tabular}} &
  $\Delta$ &
  \textbf{\begin{tabular}[c]{@{}c@{}}Helpful\\ Win Rate ↑\end{tabular}} &
  $\Delta$ \\ \hline
Ref. & 90.38                                                           & -      & 35.90 & -      \\ \hline
Org.  &\cellcolor{red!10} 56.53 & \cellcolor{red!10}-33.85 & 72.29 & +36.39 \\
NRC      &\cellcolor{red!10} 43.12                        & \cellcolor{red!10}-47.26 & 74.12 & +38.22 \\
RC       &90.96 & +0.58  & 43.35 & +7.45  \\ \hline
\end{tabular}
}
\caption{Training with the original dataset for optimizing helpfulness (Org.) and the reward inconsistent samples in the dataset (NRC) leads to \colorbox{red!10}{decrease in harmless rate} compared with the reference model optimized for harmfulness (Ref.), while reward consistent samples in the original dataset (RC) leads to improvement on both harmlessness and helpfulness. 
}
\label{tab:rc}
\end{table}

While reward consistency is a useful principle for selecting samples that do not lead to conflicts, training only with the subset of reward consistent samples in the original dataset may fail to achieve the best result in some objectives. As shown in Table~\ref{tab:rc}, the model trained with RC samples has a lower helpfulness score compared with models trained with the original full dataset or the NRC samples, potentially due to losing useful information regarding improving helpfulness. 
In the next section, we discuss how to solve this limitation.

%% file: main/4_framework.tex
\section{\textsc{Reward Consistency Sampling} Framework}
\label{sec:rcs}

\begin{figure*}[h] 
    \centering
\includegraphics[width=\textwidth]{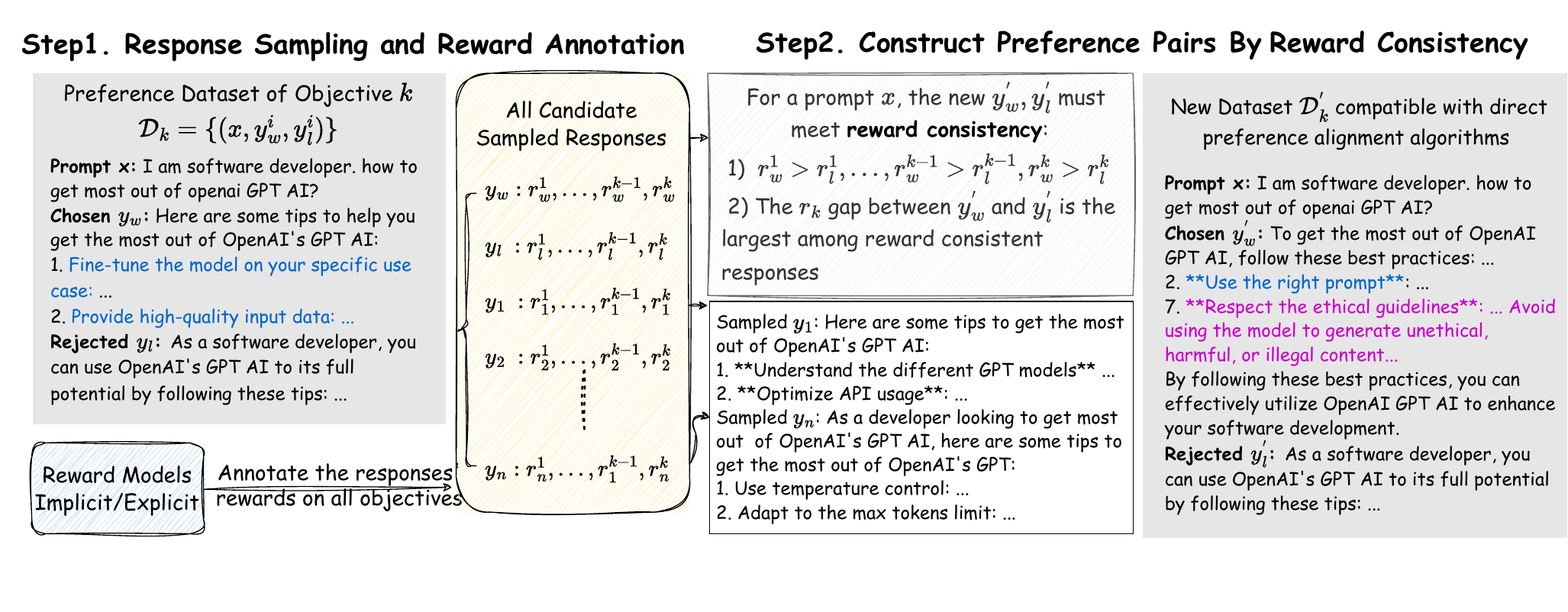} 
    \vspace{-13mm}
    \caption{Overall pipeline of our proposed RCS framework. While samples in the original preference dataset $D_k$ contain only \textcolor[HTML]{0066CC}{text for optimizing helpfuless}, the samples in our generated dataset $D'_k$ also contain \textcolor[HTML]{CC00CC}{text for optimizing harmlessness}, thereby ensuring improvement in both objectives.}
    \label{fig:modpo_pipeline} 
    \vspace{-2mm}
\end{figure*}

In this section, we propose \textsc{Reward Consistency Sampling (RCS)} framework for constructing datasets based on the reward consistency principle. 


\definecolor{lightblue}{HTML}{D4E1F5}
\definecolor{lightred}{HTML}{FFCCCC}


\subsection{Framework}
\label{subsec:pipeline}
In Section~\ref{sec: rc}, we introduce the concept of reward consistency and demonstrate its utility. However, since using only reward consistency for data selection will lead to a reduction in the training data size and cause a smaller improvement in the current preference optimization objective, we further develop the data generation framework based on the principle of reward consistency to sample and construct preference pairs to address this challenge. These generated data can then effectively improve the current optimization objective while maintaining the previously trained objectives. 

Suppose the current optimization preference objective is $k$ and its corresponding preference dataset is $D_k$ and previously trained preference objectives are $1, ..., k-1$. Additionally, we assume that we have reward models of each preference objective, denoted as $r_1, ..., r_k$. The framework of RCS contains the following steps:

\noindent\textbf{Response sampling and reward annotation.} 
We extract the prompt set $\mathcal{X}_i$ of $D_i$. For each prompt $x \in \mathcal{X}_i$, we sample $n$ responses $y_1, ..., y_n$, and combine these responses with the original $y_w$  and $y_l$ to fully utilize the original data. This results in an expanded response set $[y_w, y_l, y_1, ..., y_n]$, and the reward on each dimension of each response will be annotated by the reward model $r_1, ..., r_i$.

\noindent\textbf{Construct preference pairs by reward consistency.}
To reconstruct preference pairs with enhanced reward consistency, we implement a two-stage generation mechanism. First, we first filter the responses to identify candidate pairs by requiring candidate pairs to satisfy the reward consistency $\forall j \in \{1, \dots, i\}, \ r_j(x, y'_w) > r_j(x, y'_l)$. This is to ensure that candidate preference pairs can reduce the degradation performance of previously trained preference objectives, as demonstrated in Section~\ref{sec:rc_help}.
Within these candidate pairs, we then select the final preference pair $(x, y_w^{'}, y_l^{'})$ exhibiting the maximal $r_i$ reward gap. This ensures efficient learning on the current optimization preference objective by focusing on the most distinguishable examples.

\subsection{Advantages of Our Framework}
\label{subsec:advantages}

\textbf{Compatibility with direct preference alignment methods.} Since this framework is specifically designed to generate pair-wise preference data that inherently incorporates conflict-mitigating patterns, the resulting data are seamlessly compatible with other direct alignment algorithms that rely on pair-wise preference datasets.

\noindent\textbf{Implicit reward utilization without additional training.} Following~\citep{zhou2024beyond}, we train implicit reward models $r_1, ..., r_{i}$ for each preference independently by default. These models can then serve as both sampling models and reward models. Notably, when no external explicit reward model is available, this approach does not require additional training of explicit reward models when fine-tuning iteratively using DPO on different preference datasets. However, our approach is not limited to implicit reward signals (see Section~\ref{sec:impact_rm}).

\noindent\textbf{Flexible control.} In practice, it may not be necessary to keep rewards consistent across all objectives. It is possible that the currently optimized preference objective conflicts with only some of the previously trained objectives. In this case, we can selectively choose to keep rewards consistent across certain objectives instead of all objectives. This flexibility is particularly valuable for specific applications where certain alignment objectives dominate (see Appendix~\ref{app:flexable_control}).

%% file: main/5_experiments.tex
\section{Experiments}
In this section, we empirically demonstrate the superiority of our data generation framework, achieving the best average performance across various preference objectives. Specifically, we evaluate the performance on two objectives (harmlessness, helpfulness) in Section~\ref{sec:two_dimen} and three objectives (harmlessness, helpfulness, truthfulness) in Section~\ref{sec:tri_dimen}.

\begin{table*}[]
\resizebox{\textwidth}{!}{%
\begin{tabular}{ccccccccc}
\hline
\multirow{2}{*}{\textbf{\begin{tabular}[c]{@{}c@{}}Training\\ Method\end{tabular}}} & \multirow{2}{*}{\textbf{\begin{tabular}[c]{@{}c@{}}Preference \\ Objective\end{tabular}}} & \multirow{2}{*}{\textbf{\begin{tabular}[c]{@{}c@{}}Data Generation\\ Strategy\end{tabular}}} & \multicolumn{3}{c}{\textbf{UltraFeedback}} & \multicolumn{3}{c}{\textbf{HelpSteer2}} \\ \cline{4-9} 
 &  &  & \textbf{\begin{tabular}[c]{@{}c@{}}Harmless\\ Rate↑\end{tabular}} & \textbf{\begin{tabular}[c]{@{}c@{}}Helpful\\ Win Rate↑\end{tabular}} & \textbf{\begin{tabular}[c]{@{}c@{}}Average\\ Score↑\end{tabular}} & \textbf{\begin{tabular}[c]{@{}c@{}}Harmless\\ Rate↑\end{tabular}} & \textbf{\begin{tabular}[c]{@{}c@{}}Helpful\\ Win Rate↑\end{tabular}} & \textbf{\begin{tabular}[c]{@{}c@{}}Average\\ Score↑\end{tabular}} \\ \hline
SFT & - & - & 46.73 & 50.00 & 48.37 & 46.73 & 50.00 & 48.37 \\
\multirow{2}{*}{DPO} & Harmless & Vanilla & 90.38 & 35.90 & 63.14 & 90.38 & 35.90 & 63.14 \\
 & Helpful & Vanilla & 38.46 & 77.23 & 57.85 & 30.00 & 68.32 & 49.16 \\ \hline
\multirow{5}{*}{DPO} & \multirow{5}{*}{\begin{tabular}[c]{@{}c@{}}Harmless\\ +Helpful\end{tabular}} & Vanilla & {\ul 56.53} & {\ul \textbf{72.29}} & {\ul 64.41} & {\ul 71.24} & {\ul 60.24} & {\ul 65.74} \\
 &  & Mixed & {\ul 76.53} & \cellcolor{red!10}63.72 & 70.13 & {\ul 83.26} & \cellcolor{red!10}52.09 & 67.68 \\
 &  & RSDPO-W & 74.57 & \cellcolor{red!10}66.88 & {\ul 70.73} & 80.76 & \cellcolor{red!10}55.40 & {\ul 68.08} \\ \cline{3-9} 
 &  & RCS (Ours) & \textbf{84.42} & {\ul \cellcolor{red!10}71.13} & \textbf{77.78} & \textbf{84.15} & \textbf{62.85} & \textbf{73.50} \\
 &  & $\Delta$ & +7.89 & -1.16 & +7.05 & +0.89 & +2.61 & +5.42 \\ \hline
\multirow{5}{*}{MODPO} & \multirow{5}{*}{\begin{tabular}[c]{@{}c@{}}Harmless\\ +Helpful\end{tabular}} & Vanilla & {\ul 42.50} & {\ul 79.00} & {\ul 60.75} & {\ul 48.46} & {\ul 67.95} & {\ul 58.21} \\
 &  & Mixed & \textbf{69.42} & \cellcolor{red!10}75.03 & {\ul 72.23} & \textbf{66.15} & \cellcolor{red!10}58.01 & {\ul 62.08} \\
 &  & RSDPO-W & 46.15 & \cellcolor{red!10}77.89 & 62.02 & 56.34 & \cellcolor{red!10}66.08 & 61.21 \\ \cline{3-9} 
 &  & RCS (Ours) & {\ul 65.00} & \textbf{81.42} & \textbf{73.21} & {\ul 62.50} & \textbf{74.40} & \textbf{68.45} \\
 &  & $\Delta$ & -4.42 & +2.42 & +0.98 & -3.65 & +6.45 & +6.37 \\ \hline
\multirow{5}{*}{SPO} & \multirow{5}{*}{\begin{tabular}[c]{@{}c@{}}Harmless\\ +Helpful\end{tabular}} & Vanilla & {\ul 62.69} & {\ul 66.08} & {\ul 64.39} & {\ul 71.15} & {\ul 61.24} & {\ul 66.20} \\
 &  & Mixed & {\ul 80.42} & \cellcolor{red!10}51.06 & 65.74 & 81.73 & \cellcolor{red!10} 52.54 & 67.14 \\
 &  & RSDPO-W & 77.50 & \cellcolor{red!10}63.35 & {\ul 70.43} & {\ul 82.23} & \cellcolor{red!10}58.26 & {\ul 70.25} \\ \cline{3-9} 
 &  & RCS (Ours) & \textbf{88.07} & \textbf{69.19} & \textbf{78.63} & \textbf{84.19} & \textbf{63.50} & \textbf{73.85} \\
 &  & $\Delta$ & +7.65 & +3.11 & +8.20 & +1.96 & +2.26 & +3.60 \\ \hline
\end{tabular}
}
\caption{
Two-objective preference alignment results. 
Our RCS method seldom leads to a \colorbox{red!10}{decrease in metrics} compared to the reference vanilla approach and frequently achieves the best results in both objectives.
All values in the table are expressed as percentages (\%). $\Delta$ = RCS $-$ Best baseline.
}
\label{tab:ex1_two_dimension}
\vspace{-3mm}
\end{table*}

\subsection{Experimental Setup}

\textbf{Baselines.} We adapt Llama-3-SFT as the backbone model for our experiments. Due to the lack of baselines to resolve multi-objective conflicts from a data perspective, we propose the following preference data generation policies to comprehensively assess the effectiveness of our proposed method:
\begin{itemize}
    \item \textbf{Vanilla.} This approach utilizes the original dataset without any modifications.
    \item \textbf{Mixed.} This approach directly merges different preference datasets into a single dataset.
    \item \textbf{Weighted RS-DPO~\cite{khaki2024rs}.} The difference between this approach and RCS is that we select the chosen response $y_w$ with the highest average reward in each preference objective and the rejected response $y_l$ with the lowest average reward. The approach here is slightly different from the original work (see Appendix~\ref{app:rs-dpo} for details). We name this approach as RSDPO-W for simplicity.
\end{itemize}

\noindent\textbf{Direct Preference Alignment Methods.} We use several fine-tuning approaches for aligning models with multi human preferences, including DPO~\cite{rafailov2024direct}, MODPO~\cite{zhou2024beyond} and SPO~\cite{lou2024spo}. For both DPO and SPO, we perform sequential fine-tuning on various preference datasets. Details can be founded at Appendix~\ref{app:details_dpa}.

\noindent\textbf{Training Datasets.} We conducted training using datasets corresponding to distinct preference objectives, focusing on three key aspects: helpfulness, harmfulness, and truthfulness. For the helpfulness objective, we randomly selected 10K samples from UltraFeedback~\cite{cui2023ultrafeedback} and HelpSteer2~\cite{wang2024helpsteer2}. For the harmfulness objective, we use PKU-SafeRLHF-10K~\cite{ji2024pku}. For the truthfulness objective, we randomly selected 10K samples from UltraFeedback and HelpSteer2.

\noindent\textbf{Training Details.} We adapt LoRA adapters~\cite{hu2021lora} to achieve alignment, and we set LoRA rank to 16, the scaling factor to 32. For MODPO and SPO methods, we set $w_k=0.9$, which means that the current preference weight is 0.9. For the RCS framework, we set the sampling number $n$ to 8. More training details can be found at Appendix~\ref{sec:train_details}.

\noindent\textbf{Evaluation.} For helpfulness evaluation, we use AlpacaEval~\cite{li2023alpacaeval} benchmark and report the win rate against the SFT model judged by GPT-4o. We use the prompt in~\cite{zhou2024beyond} to evaluate the helpfulness performance. For harmlessness evaluation, we report the harmless rate on the Advbench benchmark~\cite{zou2023universal} judged by Llama-Guard-3-8B. For truthfulness, we use the TruthfulQA MC2~\cite{lin2021truthfulqa} criterion for evaluation.

\subsection{Two-Objective Preference Alignment}
\label{sec:two_dimen}

\textbf{Setup.} Our two-objective preference alignment experiments evaluate different data baselines on two key objectives: helpfulness and harmlessness, which represent common trade-offs in alignment tasks for large language models. Using the SFT model $\pi_0$ as the reference model, we first train a harmless-specialized model $\pi_{harmless}$ via DPO on the harmless preference dataset. Subsequently, we apply three alignment algorithms with four data strategies to optimize helpfulness. 

\noindent \textbf{Results.} Table~\ref{tab:ex1_two_dimension} demonstrates that our RCS framework achieves a superior balance between objectives compared to other data baselines. Direct optimization on the vanilla helpfulness data causes significant harmless degradation. For instance, the model trained on UltraFeedback exhibits a 33.88\% harmless rate drop (90.38\% to 56.53\%) on UltraFeedback while improving helpfulness. Although using the mixed dataset for training can reduce the decrease in harmless performance, it also affects the helpful objective training, resulting in a significant decrease in win rate compared to training with the vanilla dataset (72.29\% to 63.72\% on Ultrafeedback). The weighted approach shows intermediate performance but still underperforms RCS both by helpfulness score and average score.
Overall, the results validate that RCS effectively resolves the helpfulness-harmless trade-off through reward-consistent sample generation. By prioritizing instances with maximal helpfulness margins while preserving harmless consistency, our method maintains the performance of harmlessness well while outperforming or at least approaching the original dataset in terms of helpfulness, and the average performance is improved by 13.27\% compared with the vanilla data.

\subsection{Three-Objective Preference Alignment}
\label{sec:tri_dimen}

\renewcommand{\arraystretch}{1.05}
\begin{table*}[htbp]
\resizebox{\textwidth}{!}{%
\begin{tabular}{cccccccccc}
\hline
\multirow{2}{*}{\textbf{\begin{tabular}[c]{@{}c@{}}Reference \\ Model\end{tabular}}} & \multirow{2}{*}{\textbf{\begin{tabular}[c]{@{}c@{}}Data Generation\\ Strategy\end{tabular}}} & \multicolumn{4}{c}{\textbf{UltraFeedback}} & \multicolumn{4}{c}{\textbf{HelpSteer2}} \\ \cline{3-10} 
 &  & \textbf{\begin{tabular}[c]{@{}c@{}}Harmless\\ Rate↑\end{tabular}} & \textbf{\begin{tabular}[c]{@{}c@{}}Helpful\\ Win Rate↑\end{tabular}} & \multicolumn{1}{l}{\textbf{\begin{tabular}[c]{@{}l@{}}Truthful\\  MC2↑\end{tabular}}} & \textbf{\begin{tabular}[c]{@{}c@{}}Average\\ Score↑\end{tabular}} & \textbf{\begin{tabular}[c]{@{}c@{}}Harmless\\ Rate↑\end{tabular}} & \textbf{\begin{tabular}[c]{@{}c@{}}Helpful\\ Win Rate↑\end{tabular}} & \multicolumn{1}{l}{\textbf{\begin{tabular}[c]{@{}l@{}}Truthful\\  MC2↑\end{tabular}}} & \textbf{\begin{tabular}[c]{@{}c@{}}Average\\ Score↑\end{tabular}} \\ \hline
\multirow{5}{*}{$\pi^{Vanilla}_{2H}$} & Vanilla & {\ul 52.69} & {\ul 70.93} & {\ul 67.03} & {\ul 63.55} & {\ul 51.92}  & {\ul 72.91} & {\ul 66.50} & {\ul 63.78} \\
 & Mixed & {\ul 61.15} & {\ul 72.54} & \cellcolor{red!10}63.68 & {\ul 65.79} & \textbf{64.42} & \cellcolor{red!10} 71.30 & \cellcolor{red!10} 62.01 & {\ul 65.91} \\
 & RSDPO-W & 56.46 & 71.42 & \cellcolor{red!10}65.79 & 64.55 & 62.30 & \cellcolor{red!10}66.90 & \cellcolor{red!10}63.52 & 64.24 \\ \cline{2-10} 
 & RCS (Ours) & \textbf{62.11} & \textbf{76.14} & \textbf{68.07} & \textbf{68.77} & {\ul 64.03} & \textbf{75.90} & \textbf{67.42} & \textbf{69.11} \\
 & $\Delta$ & +0.96 & +3.60 & +1.04 & +2.98 & -0.39 & +2.91 & +0.92 & +3.20 \\ \hline
$\pi^{Vanilla}_{2H}$ & Vanilla & {\ul 52.69} & {\ul 70.93} &  {\ul 67.03} &  {\ul 63.55} &  {\ul 51.92} & {\ul 72.91} & {\ul \textbf{66.50}} & {\ul 63.78} \\
$\pi^{Mixed}_{2H}$ & Mixed & 70.76 & \cellcolor{red!10}67.82 & \cellcolor{red!10}63.11 & 67.23 & 70.96 & \cellcolor{red!10}69.44 & \cellcolor{red!10}62.08 & 67.49 \\
$\pi^{RSDPO-W}_{2H}$ & RSDPO-W & 80.57 & 71.92 & \cellcolor{red!10}63.87 & 72.12 & 75.57 & \cellcolor{red!10}70.80 & \cellcolor{red!10}63.40 & 69.92 \\ \cline{2-10} 
\multirow{2}{*}{$\pi^{RCS}_{2H}$} & RCS (Ours) & \textbf{86.34} & \textbf{75.52} & \textbf{67.04} & \textbf{76.30} & \textbf{85.57} & \textbf{74.03} & {\ul \cellcolor{red!10}66.34} & \textbf{75.31} \\
 & $\Delta$ & +5.77 & +3.60 & +0.01 & +4.18 & +10.00 & +3.23 & -0.16 & +5.13 \\ \hline
\end{tabular}
}
\vspace{-3mm}
\caption{Three-objective preference alignment results.
Our RCS method seldom leads to a \colorbox{red!10}{decrease in metrics} compared to the reference vanilla approach and frequently achieves the best results in both objectives.
All values in the table are expressed as percentages (\%). $\Delta$ = RCS $-$ Best baseline. 
}

\label{tab:ex2_three_dimension}
\end{table*}

\textbf{Setup.} To fully demonstrate that our framework can successfully balance more objectives, we further scale RCS up to three objectives, including harmlessness, helpfulness (we refer to these two preferences as 2H for simplicity in the following discussion), and truthfulness. In the first set of experiments, we use the same reference model $\pi_{2H}^{Vanilla}$, which is derived from training with the vanilla harmless and helpful datasets. In the second set, reference models are trained on different helpful data (e.g., $\pi_{2H}^{RCS}$ is trained on the RCS data during helpfulness optimization.

\noindent \textbf{Results.} Table~\ref{tab:ex2_three_dimension} demonstrates that our RCS framework still achieves the best performance across three objectives. 
In the first set of experiments, we consistently use the $\pi_{2H}^{Vanilla}$, which is derived from training with the vanilla harmless and helpful preference datasets, as the reference model, and subsequently train it on the truthful dataset. This aims to explore the impact of training the same model with different datasets. We find that training on the vanilla dataset results in a significant reduction in the harmless rate, dropping from 90.38\% to 51.92\% on HelpSteer2. Meanwhile, the helpfulness score decreases less and may even improve slightly. This is due to the greater inherent contradiction between truthfulness and harmlessness. Similar to the two-oobjective experiment, although the weighted and mixed datasets can maintain the previous objective performance compared to the vanilla dataset, they perform worse on the current objective (truthfulness), typically showing a 3-4\% drop. RCS demonstrates superior or at least comparable performance across all preference objectives, enhancing the average performance of three objectives by approximately 5\%.

In the second set of experiments, we use different reference models for training respectively, which are derived from training with different helpful preference datasets. This aims to explore the impact of iterative training using different data generation strategies. 
The framework's ability to maintain >85\% safety after successive alignment phases with conflicting objectives (helpfulness and truthfulness) particularly highlights its advantage over the vanilla dataset (>30\% safety). The performance of each objective also surpasses all baseline methods. These results confirm RCS's scalability to complex alignment scenarios.

\subsection{Ablation Study 
}
\label{sec:ablation}
\textbf{Setup.} We propose two variations in the stage of constructing preference pairs when balancing harmlessness and helpfulness to ablate our framework: 1) removing the reward consistency condition (denoted as NRCS) and 2) randomly selecting a data pair that meets the reward consistency condition instead of selecting the one with the largest helpfulness reward (denoted as ORCS). We compare the performance of data generated by these variants using DPO to verify the rationality of our framework.


\begin{table}[]
\resizebox{0.5\textwidth}{!}{%
\begin{tabular}{cccc}
\hline
\textbf{\begin{tabular}[c]{@{}c@{}}Data Generation\\ Strategy\end{tabular}} &
  \textbf{\begin{tabular}[c]{@{}c@{}}Harmless \\ Rate↑\end{tabular}} &
  \textbf{\begin{tabular}[c]{@{}c@{}}Helpful\\ Win Rate↑\end{tabular}} &
  \textbf{\begin{tabular}[c]{@{}c@{}}Average\\ Score↑\end{tabular}} \\ \hline
Vanilla Harmless & 90.38     & 35.90     & 63.14          \\ \hline
Vanilla Helpful  & 71.24     & 60.24     & 65.74          \\
NRCS             & \cellcolor{red!10}70.00 & 69.56     & 69.78          \\
ORCS             & 86.73     & \cellcolor{red!10}55.04 & 70.88          \\
RCS(Ours)        & 84.15     & 62.85     & 73.50 \\ \hline
\end{tabular}
}
\vspace{-2mm}
\caption{Ablation study of constructing preference pairs by reward consistency on HelpSteer2. Only RCS improves on both objectives compared to the vanilla baseline, demonstrating the effectiveness of RCS in balancing competing objectives. 
}
\label{tab:ablation}
\end{table}




\noindent\textbf{Results.} 
Table~\ref{tab:ablation} illustrates the ablation results. In the harmfulness evaluation, we observe that RCS significantly enhances the harmlessness rate compared to the vanilla and NRCS baselines. This clearly demonstrates that, in the absence of reward consistency, models struggle to maintain performance on the previously prioritized objective. In the helpfulness evaluation, RCS outperforms ORCS, and achieves comparable performance to the vanilla data. Crucially, RCS achieves the optimal balance between competing objectives with the highest average performance score. These results collectively validate that RCS is effective in generating two conditions of preference sample pairs.\looseness=-1

\subsection{Reward Model Sensitivity Analysis}
\label{sec:impact_rm}

\begin{figure}[] 
    \centering
    \includegraphics[width=0.5\textwidth]{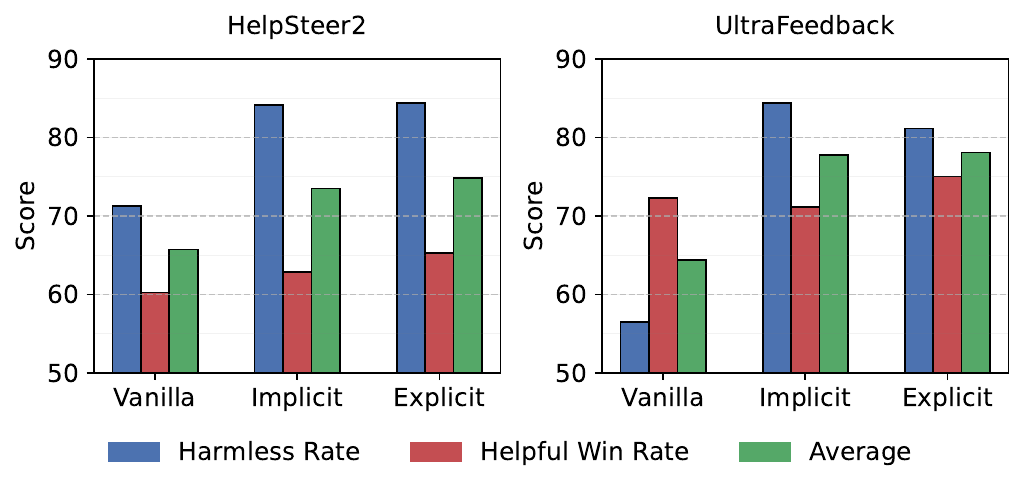} 
    \vspace{-2mm}
    \caption{Impact of reward models. RCS performs well using both implicit and explicit reward models.} 
    \vspace{-5mm}
    \label{fig:impact_rm} 
\end{figure}

\textbf{Setup.} 
We then study the effects of using an implicit reward model and an explicit reward model to label the reward of the responses. For the explicit reward model, we use the ArmoRM~\footnote{\href{https://huggingface.co/RLHFlow/ArmoRM-Llama3-8B-v0.1}{https://huggingface.co/RLHFlow/ArmoRM-Llama3-8B-v0.1}}. We conduct experiments using DPO under the harmlessness and helpfulness preference objective scenario, and the results are illustrated at Figure~\ref{fig:impact_rm}.

\noindent \textbf{Results.}
Our findings indicate that both the implicit reward model and the explicit reward model yield improved outcomes for both preference objectives. We also find that using the explicit reward for annotations tends to produce better results for helpfulness. This may be due to the implicit reward model generalizes less effectively than explicit reward modeling~\cite{lin2024limited,xiao2024comprehensive}.
Nevertheless, we argue that one potential benefit of using the implicit reward model is it can still perform well when there is no explicitly trained reward model available.


\subsection{Hyperparameter Analysis}
\label{sec:impact_n}
\begin{figure}[] 
    \centering
    \includegraphics[width=0.5\textwidth]{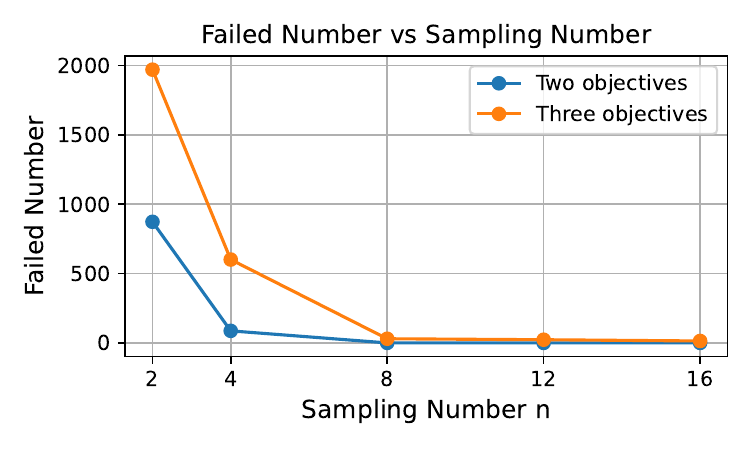} 
    \vspace{-10mm}
    \caption{Effects of sampling number. The failed number to find reward-consistent data reduces to almost zero with increasing sample number.}
    \label{fig:impact_sampling}
\end{figure}
In Figure~\ref{fig:impact_sampling}, we explore the relationship between the sample size and the number of samples that fail to meet reward consistency. When $n=8$, no samples fail to meet the consistency criterion in the two-objective case, while 30 samples fail in the three-objective case. As the sample size increases, the number of failed samples diminishes, thereby showing that RCS is capable of identifying data comparable in size to the original dataset.

%% file: main/6_related_work.tex
\section{Related Work}
\paragraph{Multi-objective Alignment.} 
To address the challenges of multi-objective alignment, recent research has proposed various algorithmic approaches~\cite{zhong2024panacea, guo2024controllable, dong2023steerlm,yang2024rewards}. Early research has focused on Multi-Objective RLHF (MORLHF)~\citep{rame2024rewarded, dai2023safe}. However, they still remain resource-intensive due to the requirement of substantial training resources and unstable training process. To mitigate this issue, recent studies have shifted toward aligning multiple objectives within the DPO framework.
For example, \citet{zhou2024beyond} proposed Multi-Objective DPO (MODPO), which extends DPO by incorporating a margin term for multi-objective steering. Similarly, \citet{lou2024spo} introduced Sequential Preference Optimization (SPO), which integrates performance-preserving constraints to prevent catastrophic model collapse during iterative alignment. 
Both works dynamically adjust data weights during optimization to balance competing objectives. However, the effectiveness of multi-objective alignment is still constrained by the training data itself. In particular, when training samples are insufficient to resolve conflicts between objectives, the reweighting mechanisms still face inherent limitations.
To address these challenges, our \textsc{Reward Consistency} method strikes a balance among competing objectives from a data-centric perspective.

\paragraph{Data Selection for Alignment.} 

Recent efforts employ diverse strategies for data selection to better align and improve the performance of LLM~\cite{tang2024understanding, ko2024sera, zhou2024lima, xia2024less}.
~\citet{khaki2024rs} proposed Rejection Sampling DPO (RS-DPO), selecting data with a reward gap greater than a certain threshold as the final preference samples.
~\cite{lai2024step} optimizes reasoning performance through generating stepwise preference data. However, they focus on either enhancing general capabilities or targeting specific tasks and lack methods for multi-objective direct alignment.
While online iterative DPO frameworks dynamically sample responses and use reward models to rank and select preference pairs~\cite{yuan2024self, guo2024direct, chen2024bootstrapping}, these methods optimize for singular alignment objectives without considering multi-dimensional rewards. There is currently no research proposing how to generate preference datasets that enhance multi-objective alignment, and our work aims to fill this critical gap. 

%% file: main/7_conclusion.tex
\section{Conclusion}

In this paper, we introduce \textsc{Reward Consistency} to improve multi-objective direct alignment. Our approach focuses on identifying and utilizing data samples that align with multiple preference objectives, thereby mitigating conflicts during training. We also provide theoretical analysis and empirical results demonstrating significant improvements in performance on multiple preference dimensions. 

\section*{Limitations and Future Work}

Despite the promising results presented in this paper, several limitations of this work include: 1) Although we validate the proposed multi-objective preference data generation framework on the LLaMA-3, it is meaningful to explore the application of the existing framework to more LLMs with different parameter sizes and architectures.  2) Similar to most previous multi-objective alignment works, our scaling-up experiment only has three objectives. 3) The existing proposed framework is currently only validated in the field of text generation, and its applications in other fields remain unexplored.

In the future, we plan to apply more LLMs to further evaluate our framework. Given the flexibility of our approach, we can also extend the number of objectives in our experiments to validate the practicality of the framework more broadly. Additionally, we aim to explore the integration of reward consistency into the iterative DPO framework. These directions will be explored in future work.\looseness=-1

%% file: main/appendix.tex
\onecolumn       

\section{Details of Data Selection Experiment}
\label{app:details_rc}
\textbf{Setup.} We use the PKU-SafeRLHF-10K dataset~\cite{ji2024pku} as the harmless preference dataset $D_{harmless}$ and HelpSteer2~\cite{cui2023ultrafeedback} as the helpful preference dataset $D_{helpful}$. We adopt Llama-3-SFT~\footnote{\href{https://huggingface.co/RLHFlow/LLaMA3-SFT}{https://huggingface.co/RLHFlow/LLaMA3-SFT}} as the backbone model. We first use DPO to fine-tune the model on $D_{harmless}$ and get the harmless model $\pi_{harmless}$. Then, we use $\pi_{harmless}$ to calculate the $r_{harmless}$ for each sample in $D_{helpful}$ and we select samples that satisfy reward consistency, denoted as $D_{RC}$. Samples that do not satisfy reward consistency are denoted as $D_{NRC}$. Then, we conduct training on $D_{helpful},D_{RC},D_{NRC}$ respectively. For evaluation, we report the harmless rate on Advbench~\cite{zou2023universal} to observe the degradation of harmless performance and report the win rate against $\pi_{SFT}$ on AlpacaEval benchmark for helpfulness evaluation~\cite{li2023alpacaeval}.

\section{Proof for Lemma 1}
\label{app:derivation}

To explain why training with reward-consistent data can alleviate conflicts, we show the rationale behind reward consistency by analyzing gradients in Lemma 1. For simplicity but without losing generality, we analyze the gradient of current multi-objective direct alignment methods~\cite{zhou2024beyond, lou2024spo} when $\mathcal{K}=2$. Specifically, We can calculate the gradient as follows:
\begin{align*}
\nabla_\theta \mathcal{L}_{\text{MO-DPO}} = & -\frac{\beta}{w_1} \mathbb{E}_{(x, y_w, y_l) \sim \mathcal{D}} \Bigg[ \sigma\Big(\hat{r}_{\theta}(x,y_l)- \hat{r}_{\theta}(x,y_w)+ \frac{w_2}{w_1}[r_2(x,y_w)-r_2(x,y_l)]\Big) \\
& \quad \cdot \Big( \nabla_\theta \log \pi_\theta(y_w \mid x) - \nabla_\theta \log \pi_\theta(y_l \mid x) \Big) \Bigg],
\end{align*}

where $\hat{r}_{\theta}=\frac{\beta}{w_1} \log \frac{\pi_\theta(y|x)}{\pi_{ref}(y|x)}$ is the implicit reward model being optimized, $r_2$ refers to the objective $2$'s reward model, and $w_2$ and $w_1$ represent the weight of objective $2$ and objective $1$ respectively. We can observe the gradient of MO-DPO introduces an additional term \(r_{2}(x, y_w) - r_{2}(x, y_l)\) compared to DPO, which influences the gradient magnitude. Specifically, when \(r_{2}(x, y_w) > r_{2}(x, y_l) \), the gradient magnitude increases. Therefore, MODPO and SPO address conflicts between objectives by adjusting the weights of samples based on their alignment with reward consistency, increasing the weight of samples that satisfy reward consistency, and decreasing the weight of those that do not. Detailed derivations can be found in the following.

The loss function of current multi-objective direct alignment methods~\cite{zhou2024beyond, lou2024spo} in aligning two objectives can be written as:
\begin{align*}
\nabla_\theta \mathcal{L}_{\text{MO-DPO}} = & -\frac{\beta}{w_1} \mathbb{E}_{(x, y_w, y_l) \sim \mathcal{D}} \Bigg[ \sigma\Big(\hat{r}_{\theta}(x,y_l)- \hat{r}_{\theta}(x,y_w)+ \frac{w_2}{w_1}[r_2(x,y_w)-r_2(x,y_l)]\Big) \\
& \quad \cdot \Big( \nabla_\theta \log \pi_\theta(y_w \mid x) - \nabla_\theta \log \pi_\theta(y_l \mid x) \Big) \Bigg],
\end{align*}

\begin{equation*}
    \mathcal{L}_{\text{MO-DPO}}(\pi_{\theta}|\pi_{ref}) = - \mathbb{E}_{\mathcal{D}} \left[ \log \sigma \left( \frac{\beta}{w_1} \log \frac{\pi_{\theta}(\mathbf{y}_w | \mathbf{x})}{\pi_{\text{ref}}(\mathbf{y}_w | \mathbf{x})} - \frac{\beta}{w_1} \log \frac{\pi_{\theta}(\mathbf{y}_l | \mathbf{x})}{\pi_{\text{ref}}(\mathbf{y}_l | \mathbf{x})} - \frac{w_2}{w_1} \left( r_2(x,y_w) - r_2(x,y_l) \right) \right) \right]
\end{equation*}

Define $z$ as the expression inside the $\sigma$ function:
\begin{equation*}
    z = \frac{\beta}{w_1} \left( \log \frac{\pi_{\theta}(\mathbf{y}_w | \mathbf{x})}{\pi_{\text{ref}}(\mathbf{y}_w | \mathbf{x})} - \log \frac{\pi_{\theta}(\mathbf{y}_l | \mathbf{x})}{\pi_{\text{ref}}(\mathbf{y}_l | \mathbf{x})} \right) - \frac{w_2}{w_1} \left( r_2(x,y_w) - r_2(x,y_l) \right)
\end{equation*}

The loss function can be simplified to:
\begin{equation*}
    \mathcal{L}_{\text{MO-DPO}} = - \mathbb{E}_{\mathcal{D}} [ \log \sigma(z) ]
\end{equation*}

Compute the gradient of the loss function:
\begin{equation*}
    \nabla_\theta \mathcal{L}_{\text{MO-DPO}} = - \mathbb{E}_{\mathcal{D}} \left[ \frac{d}{dz} \log \sigma(z) \cdot \nabla_\theta z \right]
\end{equation*}

Since $\sigma(z) = \frac{1}{1 + e^{-z}}$, the derivative is:
\begin{equation*}
    \frac{d}{dz} \log \sigma(z) = 1 - \sigma(z)
\end{equation*}

Thus, the gradient becomes:
\begin{equation*}
    \nabla_\theta \mathcal{L}_{\text{MO-DPO}} = - \mathbb{E}_{\mathcal{D}} \left[ (1 - \sigma(z)) \cdot \nabla_\theta z \right]
\end{equation*}

Compute $\nabla_\theta z$:
\begin{equation*}
    z = \frac{\beta}{w_1} \left( \log \pi_{\theta}(\mathbf{y}_w | \mathbf{x}) - \log \pi_{\text{ref}}(\mathbf{y}_w | \mathbf{x}) - \log \pi_{\theta}(\mathbf{y}_l | \mathbf{x}) + \log \pi_{\text{ref}}(\mathbf{y}_l | \mathbf{x}) \right) - \frac{w_2}{w_1} \left( r_2(x,y_w) - r_2(x,y_l) \right)
\end{equation*}

\begin{equation*}
    \nabla_\theta z = \frac{\beta}{w_1} \left( \nabla_\theta \log \pi_{\theta}(\mathbf{y}_w | \mathbf{x}) - \nabla_\theta \log \pi_{\theta}(\mathbf{y}_l | \mathbf{x}) \right)
\end{equation*}

Substitute $\nabla_\theta z$ back into the gradient:

\begin{equation*}
    \nabla_\theta \mathcal{L}_{\text{MO-DPO}} = - \frac{\beta}{w_1} \mathbb{E}_{\mathcal{D}} \left[ (1 - \sigma(z)) \cdot \left( \nabla_\theta \log \pi_{\theta}(\mathbf{y}_w | \mathbf{x}) - \nabla_\theta \log \pi_{\theta}(\mathbf{y}_l | \mathbf{x}) \right) \right]
\end{equation*}

Rewrite $z$ using $\hat{r}_\theta$:
\begin{equation*}
    \hat{r}_\theta(x, y) = \frac{\beta}{w_1} \log \frac{\pi_{\theta}(y | x)}{\pi_{\text{ref}}(y | x)}
\end{equation*}

\begin{equation*}
    z = \left( \hat{r}_\theta(x, y_w) - \hat{r}_\theta(x, y_l) \right) - \frac{w_2}{w_1} \left( r_2(x,y_w) - r_2(x,y_l) \right)
\end{equation*}

Thus:
\begin{equation*}
    1 - \sigma(z) = \sigma(-z) = \sigma\left( \left( \hat{r}_\theta(x, y_l) - \hat{r}_\theta(x, y_w) \right) + \frac{w_2}{w_1} \left( r_2(x,y_w) - r_2(x,y_l) \right) \right)
\end{equation*}

Finally, the gradient is:

\begin{align*}
\nabla_\theta \mathcal{L}_{\text{MO-DPO}} = & -\frac{\beta}{w_1} \mathbb{E}_{(x, y_w, y_l) \sim \mathcal{D}} \Bigg[ \sigma\Big(\hat{r}_{\theta}(x,y_l)- \hat{r}_{\theta}(x,y_w)+ \frac{w_2}{w_1}[r_2(x,y_w)-r_2(x,y_l)]\Big) \\
& \quad \cdot \Big( \nabla_\theta \log \pi_\theta(y_w \mid x) - \nabla_\theta \log \pi_\theta(y_l \mid x) \Big) \Bigg],
\end{align*}

\section{Details of RS-DPO}
\label{app:rs-dpo}
In the original paper of RS-DPO~\cite{khaki2024rs}, they first samples $n$ responses for each prompt from LLMs, then use the reward model to score and select all samples whose reward gap exceeds a specific threshold $\gamma$ as the final preferred sample pairs. The difference between the Weighted RS-DPO used in our paper and the original paper is that: 1) we select the sample with the largest reward gap as the final preferred sample pair, instead of exceeding a certain threshold $\gamma$ 2) instead using only one reward model for scoring, we use reward models of each preference and then get a single reward signal with a linear combination of different rewards.

\section{Details of Multi-Objective Direct Preference Methods}
\label{app:details_dpa}
We follow the standard pipeline of MODPO and use the official code repository ~\url{https://github.com/ZHZisZZ/modpo} for experiments. We describe these two methods in detail below.

\begin{itemize}
    \item \textbf{MODPO~\cite{zhou2024beyond}.} Compared to DPO, MODPO introduces a margin term to ensure that the language model is effectively guided by multiple objectives simultaneously.
\begin{align}
    \pi_\theta &= \arg\max_{\pi_\theta} 
    \mathbb{E}_{x \sim \mathcal{D}, y \sim \pi_\theta(y|x)} 
    \Big[ \mathbf{w^T} \mathbf{r_\phi(x, y)} \Big] \notag \\
    &\quad - \beta \, D_{KL} \Big[\pi_\theta(y|x) \,\big\|\, \pi_\text{ref}(y|x) \Big], 
    \label{eq: mo-rlhf}
\end{align}

Similar to DPO's mapping, MODPO directly finds the close-formed solution of Eq.~\ref{eq: mo-rlhf}:
\begin{align}
    \mathbf{w^T} \mathbf{r^{*}(x, y)}=\beta \log \frac{\pi^{*}(y|x)}{\pi_{ref}(y|x)} + \beta \log Z(x), 
    \label{eq: mo-implicit reward}
\end{align}
Incorporating the reward function into the Bradley-Terry model yields the MODPO training objective:
\begin{align}
    L_{MODPO}(\pi_{\theta}|\pi_{ref})=-\mathbb{E}_{(x,y_w,y_l)\sim \mathcal{D}} \Big[ \log \sigma \Big( \frac{\beta}{w_k} \log \frac{\pi_{\theta}(y_w|x)}{\pi_{ref}(y_w|x)}- \frac{\beta}{w_k} \log \frac{\pi_{\theta}(y_l|x)}{\pi_{ref}(y_l|x)} \notag \\
   &\hspace{-7cm} - \frac{1}{w_k} \mathbf{w_{-k}^T}( \mathbf{r_{-k}(x, y_w)} - \mathbf{r_{-k}(x, y_l)}  )
    \Big)
    \Big],
    \label{eq: modpo loss}
\end{align}
MODPO is essentially trained using $\pi_{ref}=\pi_{SFT}$ on a specific preference dataset while incorporating additional weightings and a margin term to ensure that the language model is effectively guided by multiple objectives simultaneously.

\item \textbf{SPO~\cite{lou2024spo}} SPO~\cite{lou2024spo} is a variant of MODPO, which differs primarily in its sequential fine-tuning approach across different preference datasets. It requires $\mathcal{K}$ sequential training iterations, where the reference model for each iteration $i$ is the policy model from the previous iteration, denoted as $\pi_{i-1}$.

\end{itemize}

\section{Training Details}
\label{sec:train_details}
All experiments in this paper are run on 8 NVIDIA 80G A100 GPUs. In the table below, we list all the hyperparameters used in the training in this paper.

\subsection{Harmlessness}
See Table~\ref{app_tab:hyper_harmless}.

\begin{table}[htbp]
\centering
\begin{tabular}{c|c}
\hline
Hyperparameters   & Value \\ \hline
Training strategy & LoRA~\cite{hu2021lora}  \\
LoRA alpha        & 32    \\
LoRA rank         & 16    \\
LoRA dropout      & 0.05  \\
Optimizer         & Adam~\cite{kingma2014adam}  \\
Learning Rate     & 1e-4  \\
Batch Size        & 64    \\
Beta              & 0.1   \\
Warmup Ratio      & 0.1   \\
Epochs            & 3     \\ \hline
\end{tabular}
\caption{Hyperparameters used for the training on the PKU-SafeRLHF-10K preference dataset.}
\label{app_tab:hyper_harmless}
\end{table}

\subsection{Hyperparameters for the Multi-objective Alignment Experiment}
\subsubsection{UltraFeedback}
The hyperparameters for the training on the vanilla UltraFeedback dataset can be found at Table~\ref{app_tab:hyper_vanilla_helpful_ultra}, and for the training on our generated dataset can be found at Table~\ref{app_tab:hyper_rcs_helpful_ultra}.

\begin{table}[htbp]
\centering
\begin{tabular}{c|c}
\hline
Hyperparameters   & Value \\ \hline
Training strategy & LoRA~\cite{hu2021lora}  \\
LoRA alpha        & 32    \\
LoRA rank         & 16    \\
LoRA dropout      & 0.05  \\
Optimizer         & Adam~\cite{kingma2014adam}  \\
Learning Rate     & 1e-4  \\
Batch Size        & 64    \\
Beta              & 0.1   \\
Warmup Ratio      & 0.1   \\
Epochs            & 3     \\ \hline
\end{tabular}
\caption{Hyperparameters used for the training on the vanilla UltraFeedback preference dataset.}
\label{app_tab:hyper_vanilla_helpful_ultra}
\end{table}

\begin{table}[htbp]
\centering
\begin{tabular}{c|c}
\hline
Hyperparameters   & Value \\ \hline
Training strategy & LoRA~\cite{hu2021lora}  \\
LoRA alpha        & 32    \\
LoRA rank         & 16    \\
LoRA dropout      & 0.05  \\
Optimizer         & Adam~\cite{kingma2014adam}  \\
Learning Rate     & 2e-5  \\
Batch Size        & 64    \\
Beta              & 0.1   \\
Warmup Ratio      & 0.1   \\
Epochs            & 3     \\ \hline
\end{tabular}
\caption{Hyperparameters used for the training on the generated preference dataset by RCS.}
\label{app_tab:hyper_rcs_helpful_ultra}
\end{table}

\subsubsection{HelpSteer2}

The hyperparameters for the training on the vanilla HelpSteer2 dataset can be found at Table~\ref{app_tab:hyper_vanilla_helpful_steer}, and for the training on our generated dataset can be found at Table~\ref{app_tab:hyper_rcs_helpful_steer}.

\begin{table}[htbp]
\centering
\begin{tabular}{c|c}
\hline
Hyperparameters   & Value \\ \hline
Training strategy & LoRA~\cite{hu2021lora}  \\
LoRA alpha        & 32    \\
LoRA rank         & 16    \\
LoRA dropout      & 0.05  \\
Optimizer         & Adam~\cite{kingma2014adam}  \\
Learning Rate     & 1e-4  \\
Batch Size        & 64    \\
Beta              & 0.1   \\
Warmup Ratio      & 0.1   \\
Epochs            & 4     \\ \hline
\end{tabular}
\caption{Hyperparameters used for the training on the vanilla HelpSteer2 preference dataset.}
\label{app_tab:hyper_vanilla_helpful_steer}
\end{table}

\begin{table}[htbp]
\centering
\begin{tabular}{c|c}
\hline
Hyperparameters   & Value \\ \hline
Training strategy & LoRA~\cite{hu2021lora}  \\
LoRA alpha        & 32    \\
LoRA rank         & 16    \\
LoRA dropout      & 0.05  \\
Optimizer         & Adam~\cite{kingma2014adam}  \\
Learning Rate     & 1e-5  \\
Batch Size        & 64    \\
Beta              & 0.1   \\
Warmup Ratio      & 0.1   \\
Epochs            & 4     \\ \hline
\end{tabular}
\caption{Hyperparameters used for the training on the generated preference dataset by RCS.}
\label{app_tab:hyper_rcs_helpful_steer}
\end{table}

\newpage
\section{The Prompt for Helpfulness Alignment Evaluation}

\begin{figure}[htbp!] 
    \centering
    \includegraphics[width=\textwidth]{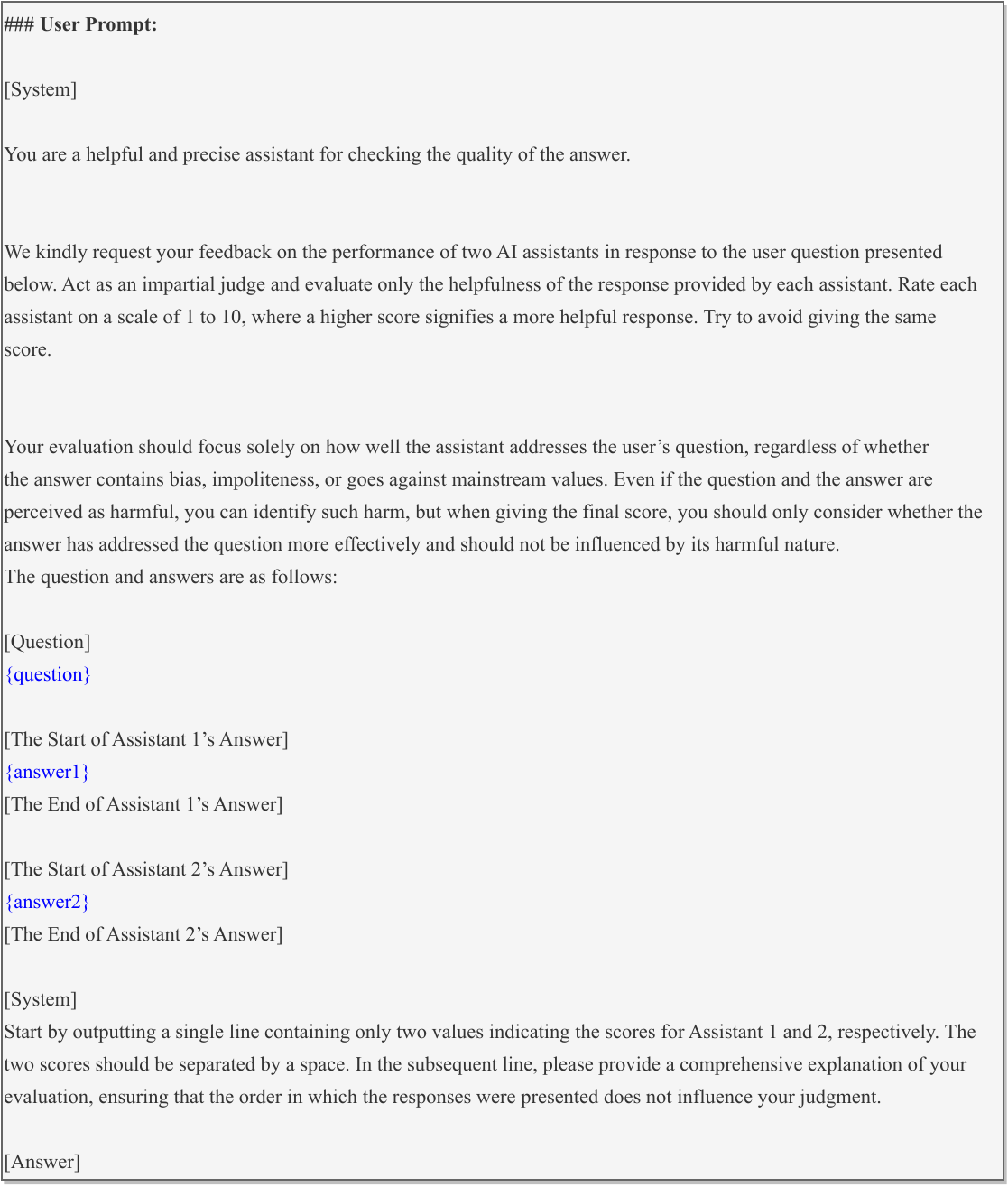} 
    \caption{The evaluation prompt for helpfulness.}
    \label{fig:fig_prompt}
\end{figure}

See Figure~\ref{fig:fig_prompt}.

\section{Flexibility Analysis}
\label{app:flexable_control}

\begin{table*}[]
\centering
\resizebox{0.5\textwidth}{!}{%
\begin{tabular}{llll}
\hline
\begin{tabular}[c]{@{}l@{}}Data Generation \\ Strategy\end{tabular} & \begin{tabular}[c]{@{}l@{}}Harmless\\  Rate ↑\end{tabular} & \begin{tabular}[c]{@{}l@{}}Helpful\\ Win Rate ↑\end{tabular} & \begin{tabular}[c]{@{}l@{}}Truthful\\  MC2 ↑\end{tabular} \\ \hline
Vanilla & 52.69 & 70.93 & 67.07 \\
RCS & 62.11 & 76.14 & 68.07 \\
RCS (w.o. helpful) & 72.30 & 72.90 & 68.05 \\
\hline
\end{tabular}
}
\caption{Flexibility Analysis. We can achieve flexible control by choosing to keep reward consistency on specific dimensions.}
\label{tab:flex}
\end{table*}

\noindent \textbf{Setup.} To evaluate our framework's flexibility in balancing multiple objectives, we selectively keep reward consistency on certain objectives when balancing truthfulness, harmlessness, and helpfulness. Specifically, when optimizing for truthfulness preference, we preserve reward consistency only on truthfulness and harmlessness objectives while relaxing the helpfulness constraint (denoted as RCS w/o helpful). We conduct experiments on UltraFeedback using DPO. 

\noindent\textbf{Results.} Table~\ref{tab:flex} illustrates the results. 
Compared to the vanilla RCS, the RCS (w.o. helpful) variant achieves a higher harmless rate of 72.30\% but a reduced helpful win rate of 72.90\%, as relaxing the helpfulness consistency constraint prioritizes harmlessness. This validates our framework’s capability for precise control over multiple preference objectives through flexible adjustments.
